\documentclass[letterpaper, 10 pt, conference]{ieeeconf}  % Comment this line out if you need a4paper

\IEEEoverridecommandlockouts                              % This command is only needed if 
\usepackage{times}
\usepackage{epsfig}
\usepackage{graphicx}
\usepackage{amsmath}
\usepackage{amssymb}
\usepackage{multirow}
\usepackage{tabularx}
\usepackage{bm}
\usepackage{xspace}
\newcommand{\bhline}[1]{\noalign{\hrule height #1}}   

\usepackage[pagebackref=true,breaklinks=true,letterpaper=true,colorlinks,bookmarks=false]{hyperref}

\title{\LARGE \bf
Audio-Visual Self-Supervised Terrain Type Discovery for Mobile Platforms
}

\author{Akiyoshi Kurobe*$^{1,2}$, Yoshikatsu Nakajima*$^{1,2}$, Hideo Saito$^{1}$ and Kris Kitani$^{2}$% <-this % stops a space
\thanks{*The authors contributed equally}% <-this % stops a space
\thanks{$^{1}$Department of Science and Technology, Keio University, Kanagawa, Japan
        {\tt\small \{kurobe, nakajima, saito\}@hvrl.ics.keio.ac.jp}}%
\thanks{$^{2}$Robotics Institute, Carnegie Mellon University, Pittsburgh, USA
{\tt\small kkitani@andrew.cmu.edu}}%
}

\begin{document}

\maketitle
\thispagestyle{empty}
\pagestyle{empty}

%%%%%%%%% ABSTRACT
\begin{abstract}
The ability to both recognize and discover terrain characteristics is an important function required for many autonomous ground robots such as social robots, assistive robots, autonomous vehicles, and ground exploration robots. Recognizing and discovering terrain characteristics is challenging because similar terrains may have very different appearances (\emph{e.g.}, carpet comes in many colors), while terrains with very similar appearance may have very different physical properties (\emph{e.g.}, mulch versus dirt). In order to address the inherent ambiguity in vision-based terrain recognition and discovery, we propose a multi-modal self-supervised learning technique that switches between audio features extracted from a mic attached to the underside of a mobile platform and image features extracted by a camera on the platform to cluster terrain types. The terrain cluster labels are then used to train an image-based convolutional neural network to predict changes in terrain types. 
Through experiments, we demonstrate that the proposed self-supervised terrain type discovery method achieves over 80\% accuracy, which greatly outperforms several baselines and suggests strong potential for assistive applications.

\end{abstract}

\section{Introduction}

Ground robots such as assistive robots (\emph{e.g.}, navigation systems for the visually impaired) and ground exploration robots are often used in open-world environments and must be able to deal with many terrain types. Therefore, the ability to automatically recognize and discover new terrain characteristics is an important function for many applications. However, it is a highly challenging task to discover terrain types robustly because similar terrains may have very different appearances (\emph{e.g.}, carpet comes in many colors), while terrains with very similar appearance may have very different physical properties (\emph{e.g.}, mulch versus dirt). 

Due to the importance of terrain recognition, many vision-based terrain classification approaches have been proposed \cite{howard2001vision, sung2010neural, mathur2012terrain, kingry2018vision}. Further, audio-based classification has been explored \cite{christie2016acoustics, valada2018deep, ojeda2006terrain, libby2012using, hadsell2013audio}.
Besides audio and visual, some researchers have made efforts to discover terrain types using vibration \cite{brooks2005vibration, ward2009speed, collins2008vibration, bai2019three} and tactile sensing \cite{takahashi2019deep, baishya2016robust}. While these existing studies have proved that each modal is effective for discovering terrain types, ambiguity remains in these methods using only a single sensing modality which may be noisy and may not be able to represent all changes in the terrain across different scenes. Therefore, we focus on an approach based on both audio and visual data, which are sensing modalities which are inexpensive, practical and easy to use.

\begin{figure}[t]
\begin{center}
   \includegraphics[width=1.0\linewidth]{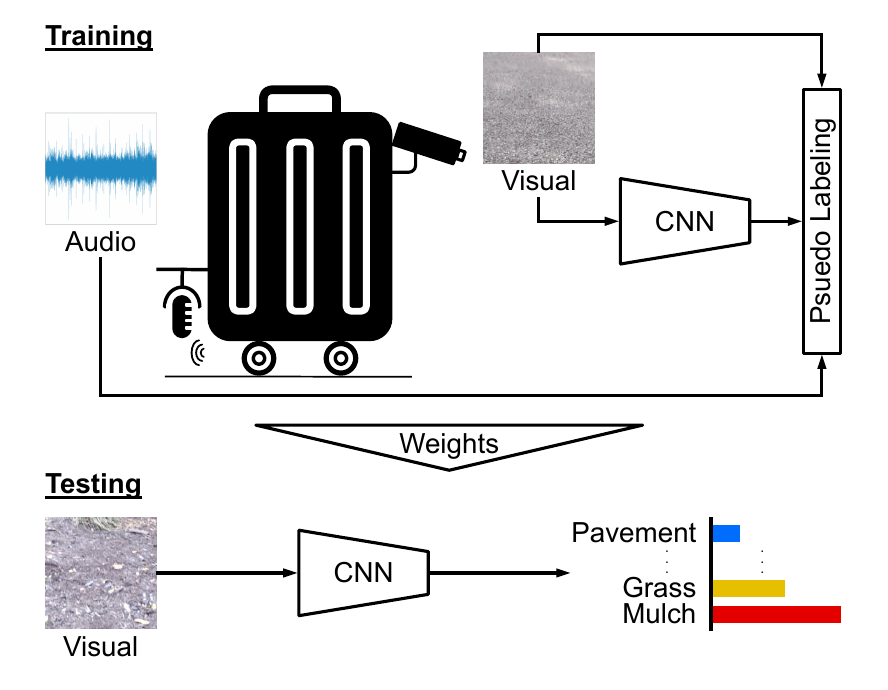}
\end{center}
   \caption{\textbf{Overview of the proposed framework}. The proposed method first generate pseudo-labels from audio recorded from a mic attached to the underside of a mobile platform and images captured by an RGB camera. These labels are utilized to train CNNs for terrain type classification in a self-supervised fashion.}
\label{fig:teaser}
\end{figure}

\begin{figure*}[t]
\begin{center}
   \includegraphics[width=0.9\linewidth]{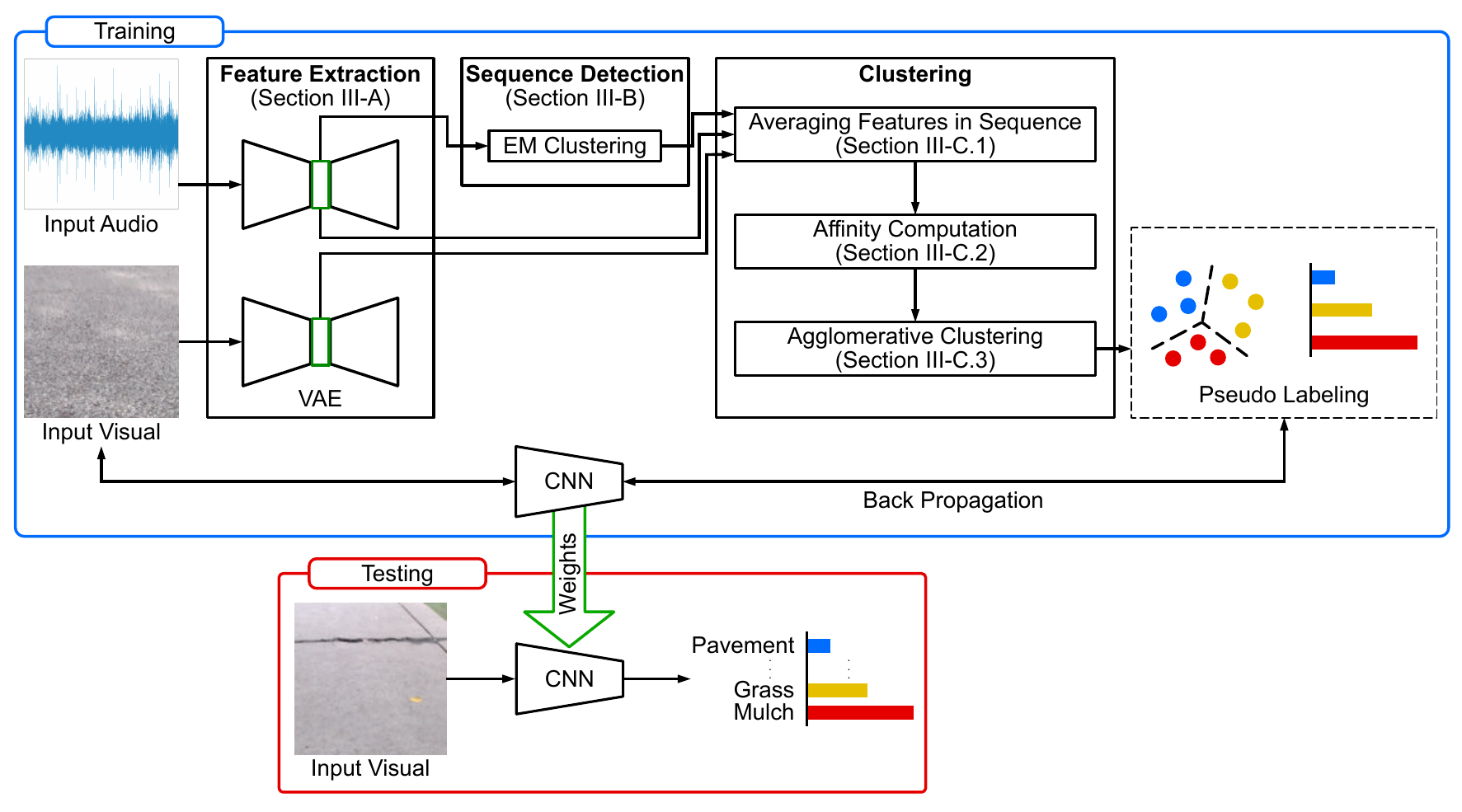}
\end{center}
   \caption{\textbf{The proposed framework}. Our adaptive multi-modal clustering approach, including sequence detection and agglomerative clustering, utilizes audio and visual cues.}
\label{fig:overview_method}
\end{figure*}

We propose a multi-modal self-supervised learning technique that switches between audio features extracted from a mic attached to the underside of a mobile platform and image features extracted by a camera on the platform to cluster terrain types. In our method, we first discover the characteristics of terrain types by audio-based clustering, which results in a discrete sequence of temporal segments. In order to reduce the noise of the features extracted over each temporal segment, \emph{e.g.} occlusions in the image or undesired environmental sounds in audio, we then compute an average features for each modality within one temporal segment. Since the temporal segments generated by the audio-based clustering tend to over segment the temporal stream of information, we implement a second phase of clustering with the averaged features to obtain temporal segments of a larger size. Since our eventual goal is the learn a vision-based terrain classifier, we use the results of the second stage of clustering to assign pseudo labels to each image in each temporal segment. These labels enable us to train an image-based convolutional neural network to predict terrain types in a self-supervised fashion (See Figure \ref{fig:teaser}).

We verify the proposed method on our own dataset, where each terrain image and audio data is associated with terrain types.
In this dataset, audio data of the friction sound is recorded with the super directional microphone heading toward the terrain and wheels. The RGB camera is mounted facing the front terrain.
This dataset is available online and would be useful for future computer vision 

%%%%%%%%% BODY TEXT

%Moreover, we release a novel dataset for indoor/outdoor mobile platforms, where each terrain image and audio data is associated with terrain types. To the best of our knowledge, this is the first dataset which contains the labeled audio data of friction sound between a wheel and the floor. Considering the recent development of assistive and navigation systems for indoor and outdoor scenes, this dataset can significantly contribute to both computer vision and robotics communities.

The contributions of this paper are as follow: 
(i) We present a self-supervised multi-modal clustering method that effectively uses the characteristics of both audio and visual cues to discover novel terrain types.
(ii) We prepare a free-to-use dataset, which contains labeled terrain images and labeled friction sounds between the terrain and the wheel.
(iii) We demonstrate the effectiveness of the proposed clustering method and framework by training and testing a CNN with several comparison approaches.

\section{Related work}\label{sec:related_work}
Research for terrain type classification has grown with the development of applications for autonomous driving and navigation systems, where some sensing modalities are utilized. In this section we describe related works in terms of terrain type discovery method, clustering method, and indoor navigation system.
\subsection{Terrain Type Discovery}
\noindent
\textbf{Vision Based.}
Howard \etal. presents a vision-based terrain classification method, where they mainly detect an edge of input images, extract a signature, and identify obstacles \cite{howard2001vision}. Sung \etal. shows that features with spatial coordinates extracted using Daub2 wavelet in the HSI color space perform well on terrain type discovery \cite{sung2010neural}. Other methods focus on analyzing terrain textures \cite{mathur2012terrain} in visual-spectrum images using Haar wavelet transforms to identify color and texture \cite{kingry2018vision}. 
The classification accuracy of vision-based terrain discovery is directly affected by appearances, although similar appearances may have very different physical properties (\emph{e.g.}, carpet versus rough concrete in Figure \ref{fig:dataset_sample}).
Considering that the field of terrain discovery is important to navigation solutions for the visually impaired, a more robust approach is desirable.
%Our assumed application includes one for visually impaired. Hence, visual only clustering approach is not robust for actual application in various situations. 

\vspace{1mm}
\noindent
\textbf{Audio Based.}
Christie \etal. presents an audio-based terrain discovering approach for legged robots by using support vector machines (SVM) on audio features which are extracted during locomotion \cite{christie2016acoustics}.
Inspired by recent developments in deep neural networks (DNNs), some methods introduce DNNs into the framework of terrain type classifications, achieving high accuracy results \cite{valada2018deep, ojeda2006terrain, libby2012using, hadsell2013audio}.
However, these methods utilize a fully-labeled dataset for training.
Thus, considering the inherent ambiguity of terrain types, these methods do not have the capability of generalizing to unseen environments.

\vspace{1mm}
\noindent
\textbf{Vibration Based.}
Vibration is often an important information source for recognizing terrain type. 
Brooks \etal. proposes vibration based classification approach, which deals with vibration data by using principal component analysis and linear discriminant analysis \cite{brooks2005vibration}.
Collins \etal. classifies terrain types using input frequency responses, which assists autonomous ground vehicle navigation \cite{collins2008vibration}.
The approach of Ward \etal. integrates vehicle speed and vibration data for training terrain type SVMs \cite{ward2009speed}.
Recently, Bai \etal proposes an approach based on 3D vibrations induced in the rover structure by the wheel-terrain interaction \cite{bai2019three}. 

\vspace{1mm}
\noindent
\textbf{LiDAR Based.}
Due to the significant role of LiDAR sensors in autonomous driving, several methods perform terrain classification with LiDAR sensors for outdoor scenes.
Vandapel \etal. and Lalond \etal. proposed a terrain classification method focusing on LiDAR point cloud segmentation \cite{vandapel2004natural, lalonde2006natural}. There are also studies that perform terrain classification by combining LiDAR point clouds and camera images \cite{laible2013terrain, laible20123d}.
Differently from these approaches, our framework works with an off-the-shelf setup (\emph{i.e.}, RGB camera and mic) and performs terrain type discovery in both indoor and outdoor scenes.

\vspace{1mm}
\noindent
\textbf{Tactile Based.}
Tactile properties such as roughness and slipperiness also represent terrain characteristics and are used in terrain classification and material estimation tasks. Baishya \etal. proposes a deep network based material estimation method which focuses on a robot finger's tactile sense \cite{baishya2016robust}. The work of Takahashi \etal. addresses the task of recognizing terrain types from visual and tactile sensors, where variational auto-encoders and recurrent neural networks are employed for feature extraction and estimation \cite{takahashi2019deep}.
As with the LiDAR based methods, these methods are expensive in terms of introducing cost for tactile sensors.
%These tactile-based approaches depend on tactile sensors, which are not very suitable for our expected scene where terrain may include some noise, such as rain or mud.

\subsection{Clustering}
For analysing features representing the target scene and captured data, clustering is a key component, and thus often applied in computer vision and robotics research.
In addition to several traditional approaches, including K-means \cite{macqueen1967some}, EM (Expectation–Maximization) clustering \cite{celeux1992classification}, and spectral clustering \cite{ng2002spectral},  
deep variational auto-encoder based clustering approach (VaDE) was proposed in recent years \cite{jiang2016variational}.
Further, their extensions for multi-source and cross-modal tasks have been proposed \cite{caron2018deep, xie2016unsupervised, radford2015unsupervised, bautista2016cliquecnn, noroozi2016unsupervised, yang2016joint, bojanowski2017unsupervised, doersch2017multi, jiang2016variational}.
Contrary to these approaches, our method switches visual- and audio-features by taking noises in terrain features into account, \emph{e.g.} human legs in images and chatting in audio.

\subsection{Indoor/Outdoor Assistive Systems}
In recent years, indoor/outdoor assistive systems have been actively developed with the improvement in depth sensors (\emph{e.g.,} Kinect and LiDAR) and global positioning systems (GPS). Kayukawa \etal. proposes a collision avoidance system for visually impaired people using both an RGB camera and a depth sensor \cite{kayukawa2019bbeep}. Terrain classification is also applied to agricultural fields for assisting agricultural tractors with LiDAR and GPS \cite{kragh2015object}. The applications of our framework would cover such indoor/outdoor assistive systems including slipping and falling avoidance.
%Hence, the proposed framework, which achieves self-supervised terrain classification with mic and RGB camera, will greatly contribute to indoor/outdoor assistive systems for avoiding slipping and falling.

% However, these approaches are not made for feature switching.

\section{Approach}
To realize self-supervised terrain type discovery, we need to perform clustering for labeling each frame (\emph{i.e.}, frames within a same cluster will be assigned the same pseudo label). A central component of our proposed approach is multi-modal clustering, where we use audio-visual cues. Figure \ref{fig:overview_method} shows an overview of the proposed framework. Given input audio and visual data, we extract features from each using a Variational Auto Encoder (VAE) (Section \ref{sec:feature_extraction}). We then perform EM (Expectation–Maximization) clustering for proposing temporal segments which have the same terrain types, \emph{i.e.} \textit{sequence} proposal (Section \ref{sec:sequence_detection}). Next, we average out noises of each feature within each \textit{sequence} (Section \ref{sec:averaged_feature}) and compute affinities between each \textit{sequence} (Section \ref{sec:calculate_connected_matrix}). Finally, we perform an agglomerative clustering based on the calculated affinities to obtain pseudo-labels for each image (Section \ref{sec:agglomerative_clustering}).

\subsection{Feature Extraction}\label{sec:feature_extraction}
In this section, we describe the details of feature extraction for both audio and visual data.
In this paper, audio and visual data represent the friction sound between the wheel and the terrain (recorded with super-directional microphone) and floor image (recorded with RGB camera), respectively.
Figure \ref{fig:setup_device} shows our setup of these sensors.

\begin{figure}[t]
\begin{center}
    \includegraphics[width=1.0\linewidth]{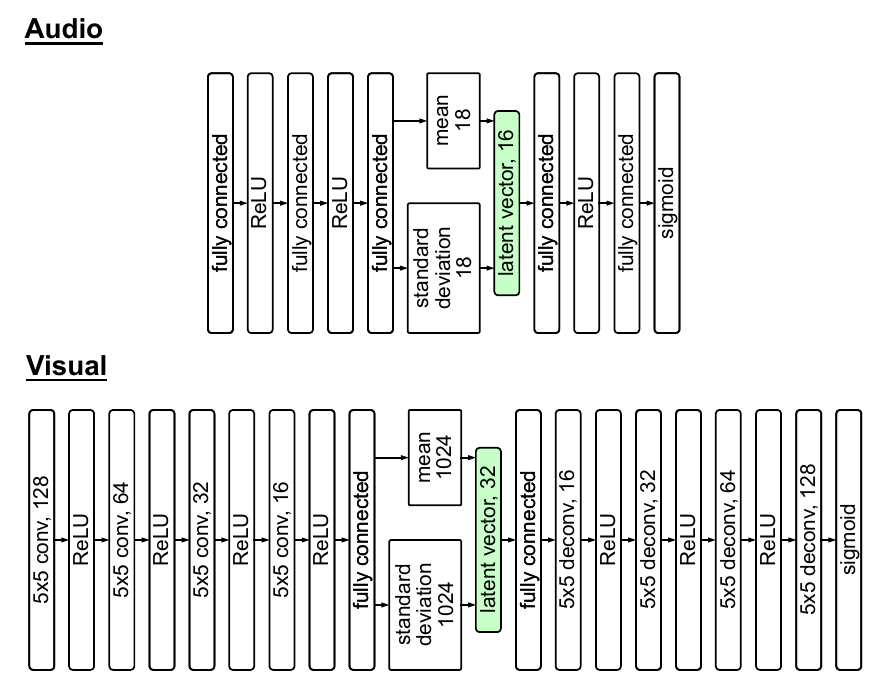}
\end{center}
   \caption{\textbf{Audio-Visual Feature Extraction}. Audio and visual feature are extracted from latent spaces of VAEs.}
\label{fig:cnn_vae}
\end{figure}

\vspace{2mm}
\noindent
\textbf{Audio.}
We set the window size for audio features long enough to being robust to undesirable noises (2.8s in experiments).
Raw audio data windows are thus too large to treat with neural networks directly, so first we compress the data.
Here, we use a simple audio feature descriptor: Mel-Frequency Cepstrum Coefficients (MFCCs) \cite{davis1980comparison}.
%MFCCs is a representation of the spectral information, which has been used extensively in a lot of fields.
We first compute 26 MFCCs, where the step between successive windows is 30 fps (frame rate of RGB camera), the length of the analysis window is 2.8 seconds, and the fast fourier transform (FFT) size is ${2^{16}}$. 
Then, we apply variational auto-encoder (VAE) feature extraction to 26 MFCCs in order to compute audio features according to a Gaussian distribution. Figure \ref{fig:cnn_vae} (upper) shows the VAE network architecture, which mainly consists of fully connected layers.
We follow the method of Kingma \etal. \cite{kingma2013auto} for training the VAE.
Through this processing, we obtain the latent vector \{$\bm{z}^{\mbox{audio}}_{t} \ | \  t\in \mathbb{Z}_{\geq 1}$ \}.

\vspace{2mm}
\noindent
\textbf{Visual.}
In order to obtain features from terrain appearances, we also extract visual latent vectors from a VAE as shown in Figure \ref{fig:cnn_vae} (lower). We resize the input image to $128 \times 128$ around the center of the image. By applying these resized images to VAE, we obtain the latent vector \{$\bm{z}^{\mbox{visual}}_{t} \ | \  t\in \mathbb{Z}_{\geq 1}$ \}. We train the VAE with the method of Kingma \etal. \cite{kingma2013auto}, as with audio features.

\subsection{Sequence Detection}\label{sec:sequence_detection}
Since clustering for all frames is noise sensitive, we perform clustering on a unit of multiple frames.
In order to propose continuous frames which have same terrain types, we perform clustering on audio features $\bm{z}^{\mbox{audio}}_{t}$.
Here, we employ EM clustering \cite{celeux1992classification}, since audio features follow a Gaussian distribution after VAE-based feature extraction.
We call a set of frames that continuously have the same clustering label $\textit{sequence}: \textbf{S}_{i}$. 
Given the clustering label \{$C_{t} \ | \ t\in \mathbb{Z}_{\geq 1}$ \} on each frame $t$, the $i$-th \textit{sequence} is defined as follows: 

\begin{equation}\label{eq:sequence}
\begin{split}
  &\textbf{S}_{i}=\left\{ t_i \leq t < t_{i+1}\ |\ t_i, t_{i+1} \in \textbf{B}\   \right\},\\
  &\textbf{B} = \left\{0,\ t_i \ |\ C_{{t_i}-1} \neq C_{{t_i}},\ t_i > 0,\  i \in \mathbb{Z}_{\geq 0} \right\}.
\end{split}
\end{equation}

Here, $\textbf{B}$ is a set of frames whose cluster changes after the frame.

\subsection{Clustering}
Although audio-based clustering has the advantage on being sensitive to the terrain changes, it tends to over-segment frames by being affected by the change of grain and tile arrangement.
The clustering method introduced in this section merges the over segmented sequences by taking advantage of visual features.

The proposed multi-modal clustering consists of the following three processes: (i) Averaging audio-visual feature in a sequence; (ii) Affinity computation between audio features and visual features; and (iii) Agglomerative clustering.
We describe the details of each processing step below.

\subsubsection{Averaging Features in Sequence}\label{sec:averaged_feature}

We first reduce external noises by averaging both audio- and visual-features within each sequence $\textbf{S}_{i}$.
This averaging further enables us to extract audio- and visual-features for each sequence $\textbf{S}_{i}$ and perform clustering in a unit of sequences, rather than frames.
We define representative features of audio $\tilde{\bm{z}}^{\mbox{audio}}_{i}$ and visual $\tilde{\bm{z}}^{\mbox{visual}}_{i}$ of the sequence $\textbf{S}_{i}$ as follows:

\begin{equation}\label{eq:average_ai}
\begin{split}
    &\tilde{\bm{z}}^{\mbox{audio}}_{i}  = \frac {1}{|\textbf{Z}^{\mbox{audio}}_{i}|} \sum_{\bm{z}^{\mbox{audio}} \in \textbf{Z}^{\mbox{audio}}_{i}} \bm{z}^{\mbox{audio}}, \\ 
    &\textbf{Z}^{\mbox{audio}}_{i}=\left\{\bm{z}^{\mbox{audio}}_{t} \ |\  t \in \textbf{S}_{i} \right\},\\
    &\tilde{\bm{z}}^{\mbox{visual}}_{i}  = \frac {1}{|\textbf{Z}^{\mbox{visual}}_{i}|} \sum_{\bm{z}^{\mbox{visual}} \in \textbf{Z}^{\mbox{visual}}_{i}} \bm{z}^{\mbox{visual}}, \\ &\textbf{Z}^{\mbox{visual}}_{i}=\left\{\bm{z}^{\mbox{visual}}_{t} \ |\  t \in \textbf{S}_{i} \right\},
\end{split}
\end{equation}

\noindent
where $\textbf{Z}^{\mbox{audio}}_{i}$ and $\textbf{Z}^{\mbox{visual}}_{i}$ denote a set of audio and visual features in $\textbf{S}_{i}$. 

\subsubsection{Affinity Computation}\label{sec:calculate_connected_matrix}

In contrast to audio features, visual features do not tend to be affected by tile arrangement changes with respect to wheel direction, since visual features depend only on their appearances.
By taking this advantage into account, our method merges these over-segmented sequences by adaptively switching clustering cues from audio to visual.

Since the noises on visual features are averaged out through the processing described in the section \ref{sec:averaged_feature}, we switch these feature spaces by simply taking the minimum value of Euclidean distance between audio- and visual-features.
The affinity between sequence $\textbf{S}_{i}$ and $\textbf{S}_{j}$ is defined as follows:

\begin{equation}\label{eq:min_each_feature}
\begin{split}
   &d(\textbf{S}_{i}, \textbf{S}_{j}) = \\
   &\min \left\{\|\tilde{\bm{z}}^{\mbox{audio}}_{i} - \tilde{\bm{z}}^{\mbox{audio}}_{j}\|_{2},\  \|\tilde{\bm{z}}^{\mbox{visual}}_{i} - \tilde{\bm{z}}^{\mbox{visual}}_{j}\|_{2} \right\}.
\end{split}
\end{equation}

\begin{figure}[t]
\begin{center}
   \includegraphics[width=0.8\linewidth]{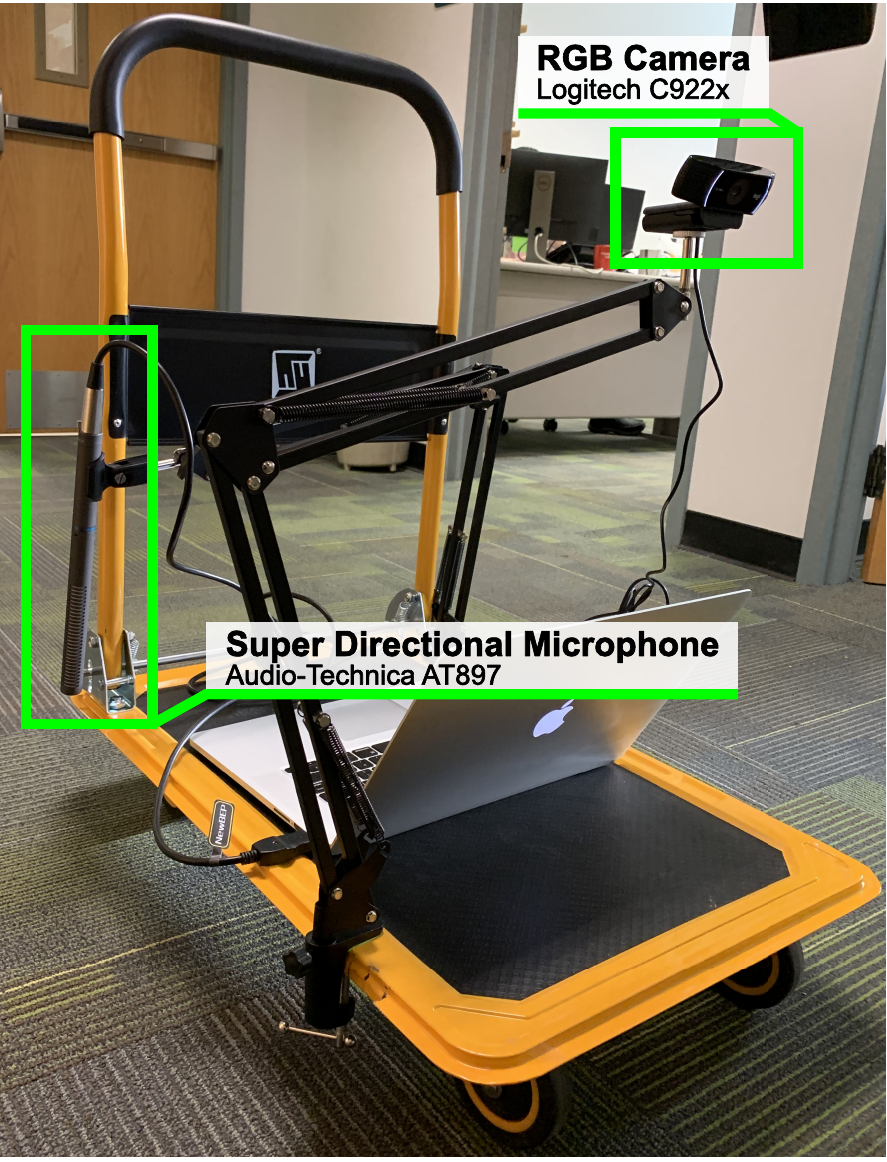}
\end{center}
   \caption{\textbf{Sensor Setup}. This figure illustrates the mounting positions of super-directional microphone and RGB camera, which are surrounded with a green square. The microphone is mounted facing the terrain and wheels in order to record the friction sound clearly. The RGB camera is mounted facing the front terrain.}
\label{fig:setup_device}
\end{figure}

\noindent
With this scheme, we are able to merge the sequences where their appearances are close enough.
Further, by considering the distance of audio features, this simple strategy is able to handle the difficulty of terrain type discovery: similar terrains may have very different appearances (\emph{e.g.}, carpet comes in many colors) but similar audio profiles.

\subsubsection{Agglomerative Clustering}\label{sec:agglomerative_clustering}
Finally, in order to obtain labels for each image, we perform agglomerative clustering on the affinity matrix whose element consists of $d(\textbf{S}_{i}, \textbf{S}_{j})$.
The clusters are directly utilized to generate pseudo labels for each sequence. 
Since the frames included in each sequence are known, we obtain labels for all frames by feeding back sequence labels to each frame.

\begin{figure}[t]
\begin{center}
   \includegraphics[width=0.98\linewidth]{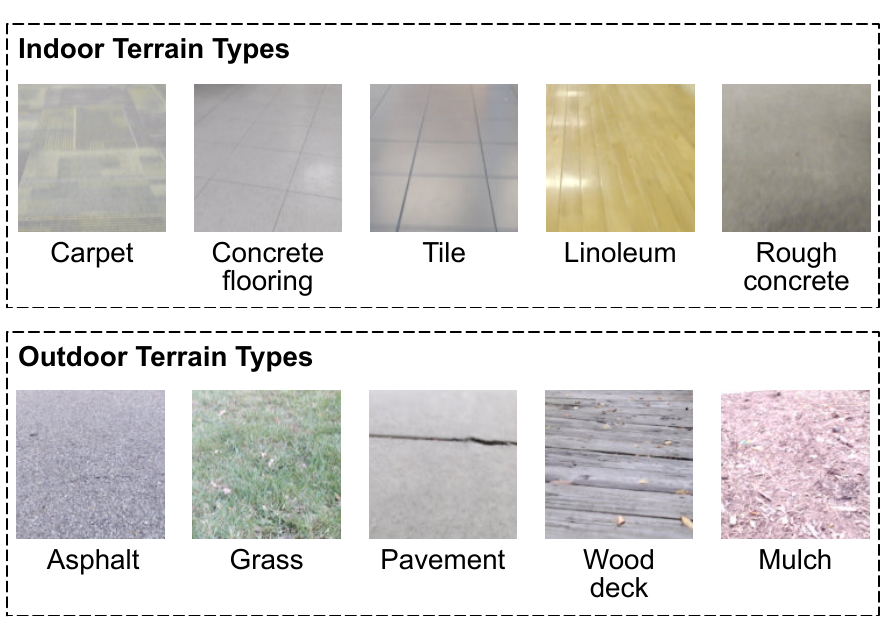}
\end{center}
   \caption{\textbf{Terrain Types}. This figure shows each terrain image example included in our dataset.}
\label{fig:dataset_sample}
\end{figure}

\section{Dataset}
In order to verify our audio-visual self-supervised terrain discovery method, we prepare a diverse terrain classification dataset for indoor/outdoor mobile platforms.
This dataset is available online and would be suitable for research of terrain type classification.
We record both audio and visual data simultaneously, where each frame is assigned to a terrain type label.
Audio data of the friction sound is recorded with the super directional microphone which is facing the terrain and wheels. Visual data is captured by the RGB camera mounted facing the front terrain.
In this section, we describe our sensor setup and the dataset structure in detail.

\subsection{Sensor Setup}
Figure \ref{fig:setup_device} shows our sensor setup. 
We put a personal computer on the dolly and connected the RGB camera and super directional microphone to it.
The sensors used are: a super directional microphone (Audio-Technica AT897 Line/Gradient Shotgun Condenser Microphone), and an RGB camera (Logitech C922x Pro Stream Webcam – Full 1080p HD Camera).
Synchronised audio-visual data is collected by scanning the scene with this dolly. 

\subsection{Dataset Detail}
Table \ref{tab:dataset} shows the detail of our dataset. 
As shown in Figure \ref{fig:dataset_sample}, there are a total ten classes of terrain types included in our dataset.
Each scene is composed of about 8000 frames, which is enough for training and testing CNNs for terrain classification.

\begin{table}[tb]
   \begin{center}
  \caption{Dataset Detail. This table shows a number of frames and terrain classes of each scene in our dataset. \label{tab:dataset}}
     \scalebox{1.0}{
       \begin{tabular}{ccccr} \bhline{1pt}
        &
        \multirow{2}{*}{No.}& \multirow{2}{*}{Scene} & \# frames & \multirow{2}{*}{Classes} \\
        {} & {} & {} & (train/test) & {} \\ \hline
		\multirow{12}{*}{\rotatebox{90}{\textbf{Indoor}}}
		&\multirow{2}{*}{1} & \multirow{2}{*}{SH}
		& \multirow{2}{*}{10694 / 7206} & Carpet \\
		&  & &  & Concrete flooring \\ \cline{2-5}
		& \multirow{3}{*}{2} & \multirow{3}{*}{NSH}
		& \multirow{3}{*}{7041 / 7698} & Tile \\
		& & {} && Carpet \\ 
		& & {} && Linoleum \\ \cline{2-5}
		&\multirow{3}{*}{3} & \multirow{3}{*}{WH}
		& \multirow{3}{*}{9046 / 8208} & Tile \\
		& & {} && Carpet \\ 
		& & {} && Linoleum \\ \cline{2-5}
	    &\multirow{4}{*}{4} & \multirow{4}{*}{GHC}
		& \multirow{4}{*}{7736 / 8397} & Tile \\
		& & {} && Carpet \\ 
		& & {} && Concrete flooring \\ 
		& & {} && Rough concrete \\ \hline
		\multirow{8}{*}{\rotatebox{90}{\textbf{Outdoor}}}
		&\multirow{3}{*}{5} & \multirow{3}{*}{Garden}
		& \multirow{3}{*}{8113 / 6543} & Asphalt \\
		& & {} && Pavement \\ 
		& & {} && Grass \\ \cline{2-5}
		&\multirow{2}{*}{6} & \multirow{2}{*}{Playground}
		& \multirow{2}{*}{3822 / 10311} & Pavement \\
		& & {} && Grass \\ \cline{2-5}
		&\multirow{3}{*}{7} & \multirow{3}{*}{Parking}
		& \multirow{3}{*}{8664 / 7093} & Pavement \\
		& & {} && Wood deck \\ 
		& & {} && Mulch \\ \hline
		\bhline{1pt}
 \end{tabular}
 }
    \end{center}
    
\end{table}

\section{Experiment}
To demonstrate the ability of the proposed method to both recognize and discover terrain types, we experiment on our dataset.
We first perform the proposed clustering method on each indoor/outdoor training scene and calculate the Normalized Mutual Information (NMI) in order to verify the effectiveness of the proposed method in comparison to other clustering approaches.
After that, we train ResNet \cite{he2016deep} using a set of input visuals linked with pseudo labels of terrain types. We then validate the trained CNN with test scenes in terms of prediction accuracy, precision, and recall values.

\subsection{Comparison Approach}
In order to verify the effectiveness of the proposed method, we experiment with comparison approaches. In this section, we verify the effectiveness (i) using multi-source (audio-visual) data; (ii) two step clustering (agglomerative clustering after sequence detection (EM clustering)); and (iii) with and without our feature switching scheme.

\vspace{2mm}
\noindent
\textbf{Single Source Clustering.}
For verifying the effectiveness of multi-source (audio-visual) data, we first test single source approaches, which directly performs EM clustering on $\bm{z}^{\mbox{audio}}_t$ and $\bm{z}^{\mbox{visual}}_t$. These comparisons reveal that single source tends to be affected by the input noise (visual-only) and over-segmentation (audio-only), compared with multi-source clustering approaches.

\vspace{2mm}
\noindent
\textbf{Multi source Clustering.}
In addition to multi-source, the proposed method employs sequence based clustering, not frame based.
Hence, we reveal the effectiveness of this processing by comparing with simple multi-source clustering, which performs EM clustering on features concatenating $\bm{z}^{\mbox{audio}}$ and $\bm{z}^{\mbox{visual}}$, which we call Audio-Visual clustering. 
Additionally, in order to verify the effectiveness of our feature switching scheme (mentioned in Section \ref{sec:calculate_connected_matrix}), we compare our method with the method of clustering on features concatenating $\tilde{\bm{z}}^{\mbox{audio}}_i$ and $\tilde{\bm{z}}^{\mbox{visual}}_i$, which does not switch feature space but uses both audio and visual. 

\vspace{2mm}
\noindent
\textbf{Deep Network Based Clustering}
As mentioned in Section \ref{sec:related_work}, deep network based clustering methods have been developed.
In our experiment, we employ a state-of-the-art deep network based clustering approach: VaDE \cite{jiang2016variational} as a representative method. 
We perform VaDE \cite{jiang2016variational} on $\bm{z}^{\mbox{audio}}_t$, $\bm{z}^{\mbox{visual}}_t$, and features concatenating $\bm{z}^{\mbox{audio}}_t$ and $\bm{z}^{\mbox{visual}}_t$.

\subsection{CNN Training.}
To evaluate the proposed framework's practicality, we train ResNet50 \cite{he2016deep} using our dataset with a pseudo labeling based on the output of the proposed clustering method for each scene.
Through our experiments, the resolution of input images is $128 \times 128$.

\begin{table}[t]
   \begin{center}
  \caption{Quantitative comparison. Single source (Audio-only  EM and Visual-only EM), multi-source (Audio-Visual RM), and deep clustering (Audio-Visual VaDE) \label{tab:quantitative_clustering} versus ours.}
     \scalebox{0.9}{
       \begin{tabular}{rccccc} \bhline{1pt}
       \multirow{2}{*}{No.}& \multirow{2}{*}{Ours} & Visual-only & Audio-only & Audio-Visual & Audio-Visual \\
        {} & {} & {EM \cite{celeux1992classification}} & {EM \cite{celeux1992classification}} & {EM \cite{celeux1992classification}} & {VaDE \cite{jiang2016variational}} \\ \hline
       1  & \textbf{88.9} & 3.1 & 82.4 & 1.8 & 5.7 \\ 
       2  & \textbf{81.9} & 14.2 & 56.6 & 14.3 & 54.2 \\
       3  & \textbf{64.9} & 12.3 & 31.7 & 10.0 & 19.3 \\
       4  & \textbf{94.3} & 36.2 & 90.1 & 48.9 & 69.1 \\
       5  & \textbf{90.7} & 36.3 & 90.7 & 63.3 & 76.8 \\
       6  & \textbf{92.2} & 48.6 & 88.6 & 83.9 & 77.2 \\
       7  & \textbf{54.1} & 21.3 & 39.7 & 30.3 & 30.4 \\ \hline
       \textbf{Total} & \textbf{81.0} & 24.6 & 68.5 & 36.1 & 50.6 \\ \bhline{1pt}

 \end{tabular}
 }
    \end{center}
    
\end{table}

\subsection{Results}
In this section, we experimentally demonstrate the performance of the proposed self-supervised multi-modal terrain type discovery method on test scenes of our dataset. In order to generate pseudo labels for training a CNN, we perform the proposed clustering method on all training scenes. After that, we train the CNN, ResNet50 \cite{he2016deep}, with the pair of pseudo labels and images, and then test on all test scenes. Through this experiment, we demonstrate performance of (i) the proposed clustering method by comparing our method with several baselines in terms of NMI; and (ii) terrain type prediction trained with the proposed framework by measuring accuracy, precision, and recall values of the trained CNN.

\begin{table}[tb]
   \begin{center}
  \caption{Ablation study on effects of clustering approaches and feature switching. \label{tab:ablation_study}}
     \scalebox{1.0}{
       \begin{tabular}{llc} \bhline{1pt}
       Feature & Clustering & Accuracy \\ \hline
       \multirow{3}{*}{Audio} & K-means \cite{macqueen1967some}  & 63.7  \\ 
       {} & EM \cite{celeux1992classification}  &  68.5 \\
       {} & VaDE \cite{jiang2016variational}  &  56.9 \\ \hline
       \multirow{3}{*}{Visual} & K-means \cite{macqueen1967some}  & 22.1  \\
       {} & EM \cite{celeux1992classification}  &  24.6 \\
       {} & VaDE \cite{jiang2016variational}  &  21.7 \\ \hline
        \multirow{3}{*}{Audio-Visual} & K-means \cite{macqueen1967some}  & 33.3  \\
       {} & EM \cite{celeux1992classification}  &  36.1 \\
       {} & VaDE \cite{jiang2016variational}  &  45.2 \\ \hline
       \multirow{2}{*}{\textbf{Ours}} & w/o feature switching (eq. \ref{eq:min_each_feature}) & 50.6  \\
       {} & w/ feature switching (eq. \ref{eq:min_each_feature}) & \textbf{81.0} \\ \bhline{1pt}
 \end{tabular}
 
 }
    \end{center}
\end{table}

\subsubsection{Clustering}
% \textcolor{blue}{(HS: Please clearly explain what you wish to say.  maybe you can explain for me in Japanese, so that I can revise this paragraph.  The next paragraph is also important, but I can not clearly understand what you wish to say.)}
We first demonstrate and analyse the performance of the proposed clustering method quantitatively and qualitatively.
For quantitative comparison, we measure NMI using the proposed training dataset.
Table \ref{tab:quantitative_clustering} and Table \ref{tab:ablation_study} show the results.
In Table \ref{tab:quantitative_clustering}, we compare the proposed method with two single source clustering approaches, where Audio-only and Visual-only features are used for EM clustering, and two multi-source clustering approaches, where Audio-Visual features are used for EM clustering and a state-of-the-art deep clustering method (VaDE).
The proposed method outperforms all comparison approaches, with an average accuracy of over 80\%.
Compared to Visual-only approach, Audio-only is able to cluster terrain more accurately, which shows that audio features are more robust to noise than visual features by setting window size long to reduce undesirable noises.
We next compare single source clustering (Visual-only and Audio-only) with multi-source clustering (Ours, Audio-Visual, and Audio-Visual VaDE).
When considering Visual-only as a criterion, the accuracy of Audio-Visual is improved, while Audio-Visual does not outperform Audio-only.
This suggests that how multi-source data is utilized for clustering is essential and verifies the effectiveness of our switching technique.
Table \ref{tab:ablation_study} shows a comparison between applied clustering algorithms, including K-means \cite{macqueen1967some}, EM \cite{celeux1992classification}, and VaDE \cite{jiang2016variational}.
The results suggest that EM clustering is superior to K-means clustering.
This is because extracted features follow a Gaussian distribution in the latent space.
In our method, we measure NMI in both our proposal (w/ feature switching) and a different approach, which concatenates $\bm{z}^{\mbox{audio}}$ and $\bm{z}^{\mbox{visual}}$ instead of switching features (w/o feature switching).
The results show that our proposed switching system greatly contributes to highly accurate clustering.
\par
\begin{figure*}[t]
\begin{center}
   \includegraphics[width=0.97\linewidth]{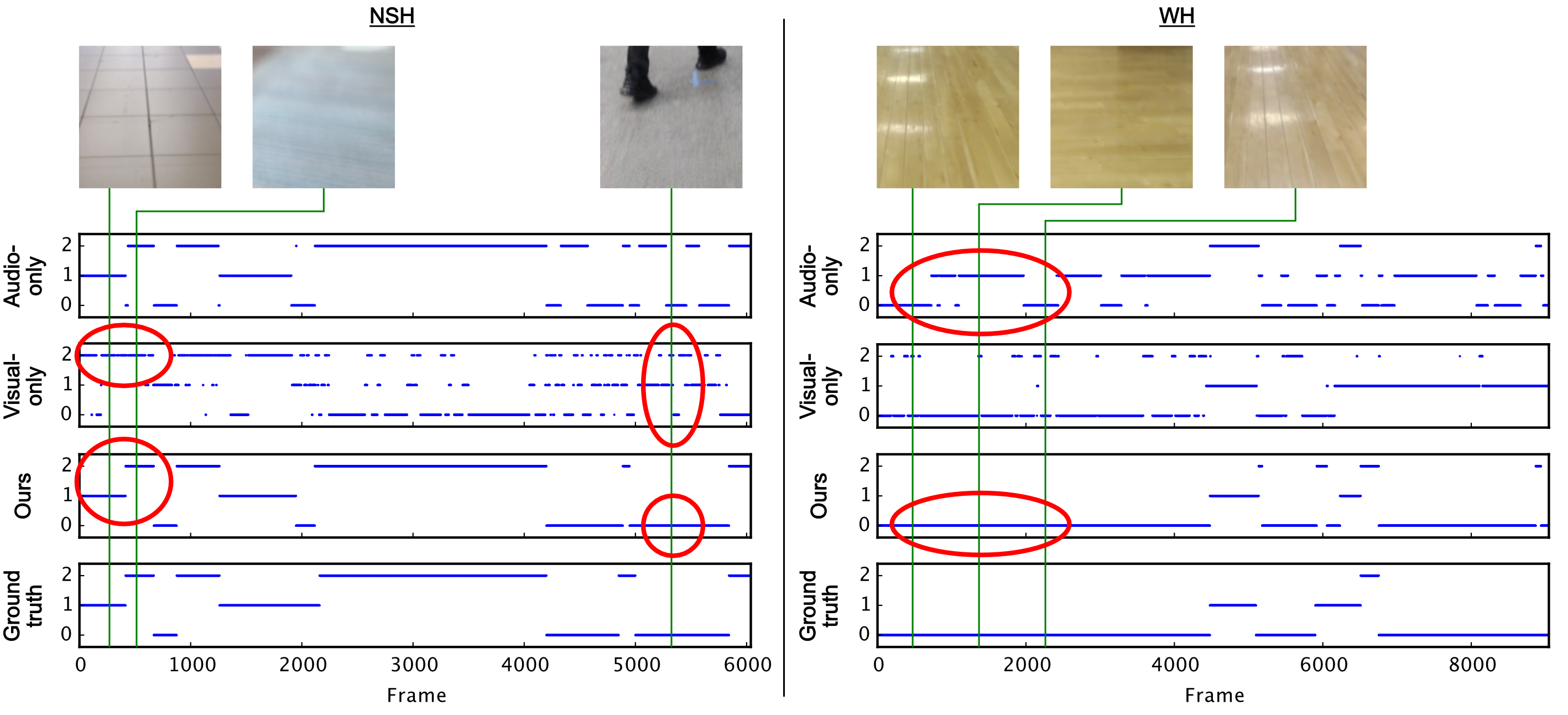}
\end{center}
   \caption{\textbf{Qualitative clustering comparison of clustering}. 
   In this comparison, we demonstrate the effectiveness of switching audio-visual features.
   In NSH scene (left), we focus on the comparison with Visual-only clustering. It tends to be affected by terrain appearance (color and texture) and noise such as human feet or wall, which is circled with red circled in figure. In WH scene (right), we focus on the comparison with Audio-only clustering, where it tends to be over-segmented when the grain and tile arrangement changes with the respect to wheel.}
\label{fig:cnn_graph}
\end{figure*}

Figure \ref{fig:cnn_graph} qualitatively shows two results of clustering on two scenes, where Audio-only, Visual-only, and Ground truth are presented.
Focusing on the red circles in the NSH scene (left), we observe that visual features are sensitive to noise (human feet) and highly dependent on terrain appearance. 
In the WH scene (right), Audio-only tends to be over-segmented because the floor grain changes with respect to the wheel (\emph{i.e.}, from vertical to parallel), while the proposed method is much accurate by switching the clustering cue to visuals.
These qualitative results verify that the proposed switching scheme is able to utilize multi-source and solve the problem of Audio-only and Visual-only approaches. 

\begin{table}[tb]
   \begin{center}
  \caption{Quantitative evaluation of terrain type predictions, in terms of precision, recall, F1-score, and accuracy. \label{tab:quantitative_cnn}}
     \scalebox{0.9}{
       \begin{tabular}{rrccccc} \bhline{1pt}
       No. & Classes & Precision & Recall & F1-score & Accuracy \\ \hline
        \multirow{2}{*}{1} & Carpet & 65.4 & 87.0 & 74.6 & \multirow{2}{*}{87.3}  \\ 
        {} & Concrete flooring & 96.1 & 87.4 & 91.5 & {} \\ \hline
        \multirow{3}{*}{2} & Tile & 80.1 & 37.7 & 51.3 & \multirow{3}{*}{74.2}  \\ 
        {} & Carpet & 88.5 & 84.3 & 86.3 & {} \\ 
        {} & Linoleum & 40.8 & 80.8 & 54.2 & {} \\ \hline
        \multirow{3}{*}{3} & Tile & 63.9 & 37.8 & 47.5 & \multirow{3}{*}{88.3}  \\ 
        {} & Carpet & 46.5 & 68.7 & 55.4 & {} \\ 
        {} & Linoleum & 92.1 & 95.7 & 93.9 & {} \\ \hline
        \multirow{4}{*}{4} & Tile & 17.0 & 27.7 & 21.1 & \multirow{4}{*}{73.6}  \\ 
        {} & Carpet & 99.6 & 71.5 & 83.2 & {} \\ 
        {} & Concrete flooring & 56.6 & 89.3 & 69.3 & {} \\  
        {} & Rough concrete & 92.8 & 68.4 & 78.7 & {} \\ \hline
        \multirow{3}{*}{5} & Asphalt & 95.5 & 89.7 & 92.5 & \multirow{3}{*}{95.7}  \\ 
        {} & Pavement & 89.8 & 98.7 & 94.1 & {} \\ 
        {} & Grass & 98.7 & 97.7 & 98.2 & {} \\ \hline
        \multirow{2}{*}{6} & Pavement & 92.5 & 98.4 & 95.6 & \multirow{2}{*}{95.5}  \\ 
        {} & Grass & 98.5 & 92.9 & 95.6 & {} \\ \hline
        \multirow{3}{*}{7} & Pavement & 91.7 & 91.0 & 91.4 & \multirow{3}{*}{89.2}  \\ 
        {} & Wood deck & 92.7 & 84.3 & 88.3 & {} \\
        {} & Mulch & 78.9 & 87.9 & 83.2 & {} \\ \bhline{1pt}

 \end{tabular}
 }
    \end{center}
\end{table}

\subsubsection{Prediction}
In Table \ref{tab:quantitative_cnn}, we present the quantitative evaluation of the terrain type prediction in terms of precision, recall, f1-score, and accuracy on the proposed test scenes.
Through all scenes our method's average accuracy is over 85\%, demonstrating the practicality of the proposed framework. 
As we experiment on both indoor/outdoor scenes, our analysis suggests that the proposed framework can be used in applications in diverse scenes.
Further, as we achieved much high accuracy (over 85\% in total), it could be argued that our framework is able to even handle delicate tasks such as assistive systems.
\par
\begin{figure*}[t]
\begin{center}
   \includegraphics[width=0.97\linewidth]{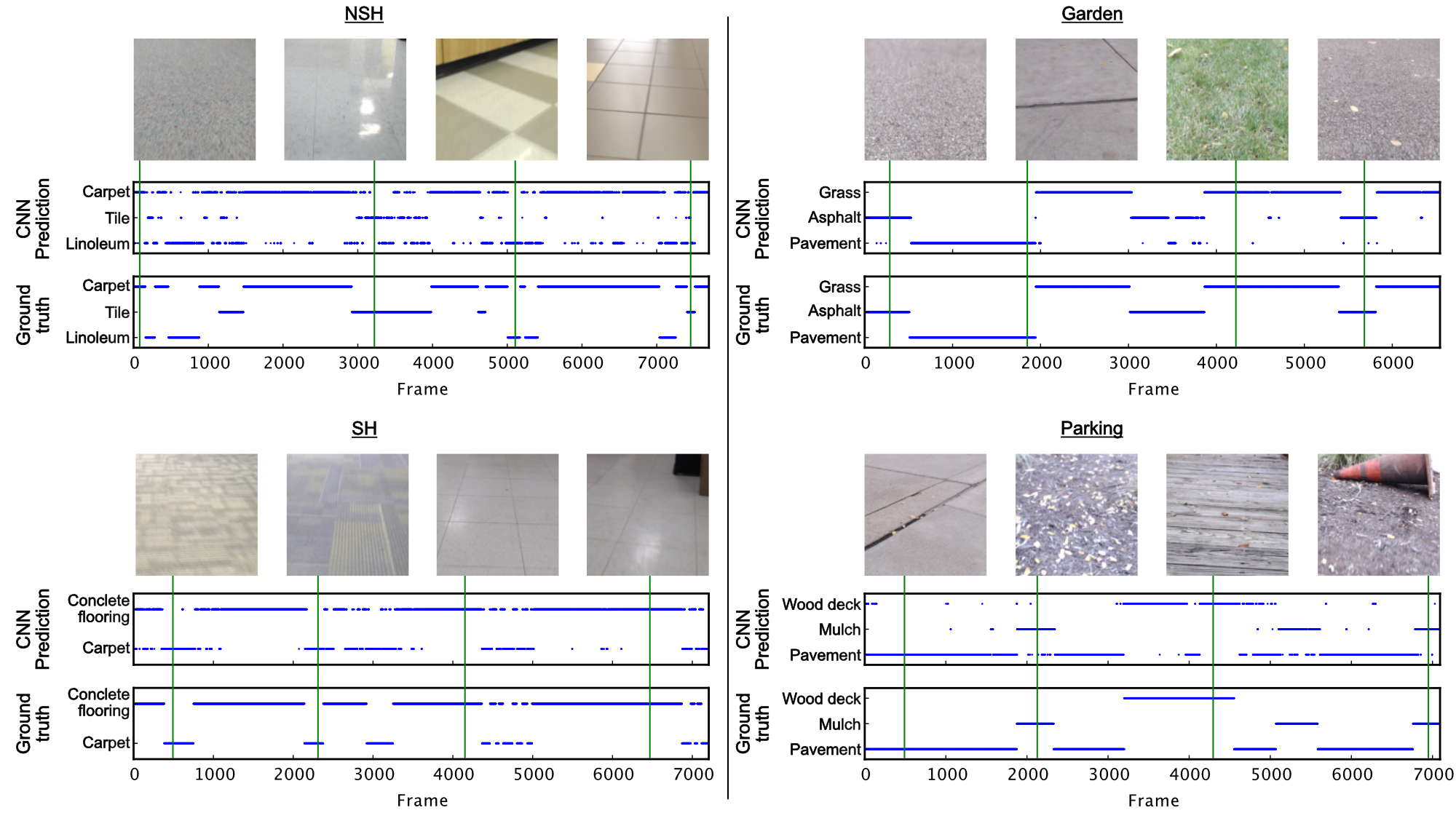}
\end{center}
   \caption{\textbf{Qualitative comparison of terrain type predictions}. The results of CNN prediction and ground truth label are visualized with blue lines. We demonstrate that the CNN correctly predicts each terrain type, although the input images have the similar color or texture. This is because pseudo labels used for training the CNN are based on adaptive switching multi-source clustering.}
\label{fig:cnn_graph2}
\end{figure*}

%Figure \ref{fig:cnn_graph2} shows qualitative results of CNN prediction with terrain images and demonstrates the practicality of the proposed framework.
%Here, the input images include the same terrain type with different colors or textures, which could not be properly recognized if the pseudo-labels are generated based on visual data alone. 
Figure \ref{fig:cnn_graph2} presents the qualitative results of CNN predictions on terrain images. Since the pseudo-labels used for CNN training are based on multi-source clustering, it is verified that terrain type can be recognized correctly even if terrain appearances are similar.

\section{Conclusion}
% Towards the estimation of the floor material for the visually impaired, we propose a method for self-supervised terrain type discovery from audio and visual data captured with a mobile platform.  Our key idea is focusing on the fact that the material characteristic appears in the friction sound between the floor and wheel.
% We demonstrate that the proposed self-supervised multi modal clustering method outperforms the baseline clustering approaches in terms of the accuracy. We also represent the framework for actual use, where in training the input is friction sound and the floor image, in testing the input is only the floor image. By training the estimated labels and testing on CNN, we verify the validity of the proposed framework. We further construct a novel dataset, which includes floor images, friction sound and each label. Hence, it verifies the effectiveness of (i) using multi-source audio-visual; (ii) two step clustering; and (iii) taking the minimum in sequence.

Towards the development of ground assistive robots, we present a novel self-supervised multi-modal terrain classification method, CNN based framework, and terrain diverse dataset. We demonstrate that the proposed clustering method is able to cluster terrain by switching between audio and visual features adaptively. Further, the practicality of the proposed framework is verified by reporting the accuracy of terrain type classification with a CNN, ResNet50, which is trained through pseudo labels generated by the proposed clustering method.

%\section{Acknowledgements}

%JST, NIDDLR

{\small
\bibliographystyle{ieee}
\bibliography{egbib}
}

\end{document}